\documentclass[letterpaper]{article}
\usepackage{iccc}
\usepackage[utf8]{inputenc}

\usepackage{times}
\usepackage{helvet}
\usepackage{courier}
\usepackage{graphicx}

\usepackage{color}

\newcommand{\citet}[1]{\citeauthor{#1} \shortcite{#1}}

\title{Modelling the Socialization of Creative Agents in a Master-Apprentice Setting: The Case of Movie Title Puns}
\author{Mika Hämäläinen\\
Department of Digital Humanities\\
Faculty of Arts\\
University of Helsinki\\
mika.hamalainen@helsinki.fi
\And
Khalid Alnajjar\\
Department of Computer Science\\
Faculty of Science\\
University of Helsinki\\
alnajjar@cs.helsinki.fi
}

\begin{document} 
\maketitle
\begin{abstract}
\begin{quote}
This paper presents work on modelling the social psychological aspect of socialization in the case of a computationally creative master-apprentice system. In each master-apprentice pair, the master, a genetic algorithm, is seen as a parent for its apprentice, which is an NMT based sequence-to-sequence model. The effect of different parenting styles on the creative output of each pair is in the focus of this study. This approach brings a novel view point to computational social creativity, which has mainly focused in the past on computationally creative agents being on a socially equal level, whereas our approach studies the phenomenon in the context of a social hierarchy.
\end{quote}
\end{abstract}

\section{Introduction}

The master-apprentice approach, as introduced by \cite{inlg}, to computational creativity has been shown to achieve creative autonomy and its creativity has been thoroughly discussed and motivated. However, the question that has remained without an answer has been the social nature of a master-apprentice pair and its effect on the creative outcome.

The approach consists of two parts: a master, which is a  genetic algorithm, and an apprentice, which is an LSTM sequence-to-sequence model. While the master is in charge of the internal appreciation of the overall system as implemented in its fitness function, the apprentice plays a crucial role in the creative autonomy as it can learn its standards partially from its master and partially from its peers.

This paper focuses on the exploration of the master-apprentice approach from a social psychological point of view. By modelling the socialization of the apprentice into a creative society consisting of the master and peers, we seek to gain a deeper understanding of the phenomenon in terms of the overall creativity of the system. In addition, modelling the social aspects of a computationally creative system can help in understanding creativity as a social phenomenon in a broader sense \cite{saunders2015computational}.

We motivate the model of socialization based on research conducted on the field of social psychology, namely developmental psychology. We select the categorization of parenting styles presented by \cite{baumrind1991influence} as the theoretical foundation of our work.

The creative task we are tackling in this paper is the creation of humorous movie titles delivering a food-related pun. This consists of taking an existing movie title such as \textit{Beauty and the Beast} and making a pun out of it such as \textit{Beauty and the Beets}. As people have been writing funny movie titles of this sort in a great abundance on the social media, we can gather parallel data easily.

\section{Related Work}

While pun generation has been vastly studied in the field of computational creativity \cite{ritchie2005computational,yu2018neural,he2019pun}, we see that the most important contribution of our paper lies in the realm of social creativity. Therefore, we dedicate this section in describing some of the practical research conducted in the computational social creativity.

Research on an agent community consisting of self-organizing maps \cite{honkela2003simulating}, although outside of the computational creativity paradigm, presents a way of simulating the emergence of language. The agents are capable of meaning negotiation and converging into a common language to communicate about edibility of different food items in their shared world.

Multi-agent systems have been studied in the context of novelty seeking in creative artifact generation \cite{linkola2016novelty}. In their study, the agents exert self-criticism and they can vote and veto on creative artifacts. Their findings suggest that multiple creative agents can reach to a higher number of novelty in their output than a single agent system.

A recent study \cite{hantula2018towards} has been conducted in social creativity in agent societies where the individuals are goal-aware. The individuals create artifacts of their own and peer up to collaborate with another agent. The agents are capable of learning a peer model that guides them in selecting a collaboration partner.

The papers discussed in this section, as well as other similar previously conducted work \cite{gabora2013meme,corneli2015implementing,pagnutti2016you}, study mostly the collaboration of agents that have an equal social status, in contrast to our case where the social status is hierarchical. Therefore we find that there's need for conducting the study presented in this paper to shed some light into asymmetrical social relations in computational creativity.

\section{Social Development}

The master-apprentice approach gives us an intriguing test bed for modelling different social interactions between the master and the apprentice. With such a complex phenomenon as human social behavior, we are bound to limit our focus on a subarea of the phenomenon. In this section, we describe different psychological approaches in understanding socialization.

Socialization, i.e. becoming a part of a social group, is an important part of the psychological development of an individual. Even to such a degree that a child who is never exposed to other people will not develop a language nor an understanding of self. Socialization, thus seems to play a crucial role in higher-level cognitive development of everything that we consider to separate a man from an animal. Perhaps this great level of importance has been the reason a great many researchers have dedicated effort in unraveling this mystery.

The ecological systems theory of social development \cite{bronfenbrenner1979ecology} highlights the importance of bidirectionality of different social groups. An individual child is in the middle of the model, but just as the immediate close family affects on the child, the child is also an actor in the process of socialization. The theory identifies multiple different systems from close family all the way to the level of the society that play a role in the social development of a child. This theory is quite complex to model computationally.

\begin{figure}[!htb]
\center{\includegraphics[width=4cm]
{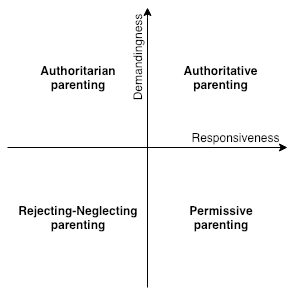}}
\caption{\label{fig:coop} Parenting styles}
\end{figure}

A take, simpler to model, on the social development is that of parenting styles \cite{baumrind1991influence}. We find these findings more suitable as a starting point for modelling the socialization of the apprentice in our master-apprentice approach. The parenting styles can be divided into four main categories: authoritative, authoritarian, permissive and rejecting-neglecting. These categories deviate from each other on the two-fold axis of demandingness and responsiveness as seen in Figure \ref{fig:coop}.

The authoritative parents are high on both demandingness and responsiveness. They set rules, but the rules are negotiable. The parenting is more supportive than punitive in nature. The authoritarian parents, on the other hand, are low on responsiveness and high on demandingness. They set non-negotiable rules and expect obedience without explanation.

The permissive parents are low on demandingess and high on responsiveness. They are very lenient and avoid confrontation. The rejecting-neglecting parents, however, are low on both axis. They hardly engage in parenting, they offer little support and do not set any rules.

\section{Creativity}

The original research on the master-apprentice approach \cite{inlg} used the creative tripod \cite{colton2008creativity} to define creativity in general and in the context of their work on creating movie titles satirical towards Saudi-Arabia by following the notions of the SPECS approach \cite{Jordanous2012}. We use the same creative tripod framework to adapt their definition into our similar task of creating movie titles with food puns. This definition provides us with a reasoned way of conducting evaluation of the overall creativity of our systems.

The creative tripod requires three key notions to be present in a system in order for it to achieve creativity. These are \textit{skill}, \textit{imagination} and \textit{appreciation}. All of these components must be present simultaneously in a creative system, or the system will lack creativity.

For our systems to exhibit skill, they will need to take a movie title as input and produce a new one with a food pun. As in the case of the earlier master-apprentice approach, the new humorous title should still communicate the original title, i.e. the original name of the movie should be recognizable.

Requirements for a pun are that it reassembles the original word in pronunciation and that it is humorous. According to \cite{elliot2003engaging}, \textit{incongruity} results in humour if it is delivered in a playful fashion and accompanied by its resolution. Another, maybe a bit more concrete way of looking at humour, is seeing incongruity as a \textit{surprise} and resolution as \textit{coherence} c.f. \cite{BROWNELL198320}.

Surprise in the context of humor means that the brain forms an expectation and this expectation is then broken by the humorous element of the pun. Such is the case in the pun \textit{Harry Potter and the Deathly Marshmallows} where the surprise is caused by the fact that the expected word \textit{Hallows} is replaced by \textit{Marshmallows}. For the pun to be coherent, it should make sense in the context of the original movie. In this case, a thought of deathly marshmallows attacking the Hogwarts, although bizarre, can still be seen as coherent.

For achieving appreciation, the systems will need to be able to assess the humorousness of the created pun in terms of surprise, coherence and sound similarity. In addition to the humour, the system should be able to evaluate the recognizability of the original title.

We define imagination by using the dichotomy of creativity introduced by \cite{boden2004creative}. This way of understanding creativity divides it in two different types: P-creativity and H-creativity. P-creativity is the minimal requirement we set for imagination of the systems, and it means that a creative entity should be able to come up with something that is novel to itself. H-creativity, on the other hand, refers to an innovation that is novel in a more global scale, i.e. nobody else has come up with a similar creative artifact before. While P-creativity is the minimum requirement, we consider H-creativity as a more desired requirement for imagination.

\section{The Data}

As our approach is to generate food related puns, we need a vocabulary consisting of food related terms. For this purpose we use the Historical Thesaurus of the Oxford English Dictionary\footnote{http://www.oed.com/thesaurus}. We use all the nouns recorded under the topic \textit{food and drink} in the \textit{external world} taxonomy. This list contains 15,314 different nouns.

We extract real movie titles from the IMDB\footnote{Dumps from https://datasets.imdbws.com/} (Internet movie database). As we want our movie title corpus to consist only of well known movies, we want to filter out all the less known indie movies. To achieve this, we filter out movies that have received less than 100,000 votes, leaving us with 1,661 movie titles. For the master and apprentices this is further limited to 1276 titles by filtering out the titles that consisted only of one word.

For parallel humorous movie title data (later peer data), we crawl comments on an Instagram post for an entertainment account\footnote{https://www.instagram.com/p/BsWki-PFbMO/}. People were encouraged to come up with creative movie titles containing a pun related to food.
The total number of comments crawled is 16,088\footnote{Crawled on the second of February}.
Then, we follow the same approach applied in~\cite{inlg} to map the crawled data to movie titles.
In summary, we preprocess the text to remove any hashtags and mentions, and then we measure the character and word edit distances between the comments and movie titles. Finally, a comment is considered to be a punny variation of the matched movie title with the least edit distance, only if it had at most three word differences while ensuring that there exist at least one word matching the movie title.
This process yields 9,294 human-authored movie titles containing a pun.

\section{The Master-Apprentice Model}

The master-apprentice model consists of a computationally creative genetic algorithm that implements the criteria set for appreciation in its fitness function and an apprentice that is an NMT (neural machine translation) model. The master generates parallel data for the apprentice to learn from, while the apprentice can also learn from its peers. In our setting, we have four different apprentices; one for each parenting style.

\subsection{Master}

\label{sec:the-master}
Inspired by the work on slogan generation presented by~\citet{alnajjar2018slogan}, we employ a similar generator to act as a master in our model.
In our case, the generator, which is a genetic algorithm, receives an original movie title as input and outputs an entire population of movie titles carrying a pun, based on the input movie title.
The master makes use of the food related vocabulary described earlier to replace words in the original title while optimizing multiple parameters to increase the aptness of the substitution and the punniness of the title.
The following subsections elucidate the algorithm.

\subsubsection{Evolutionary algorithm}
The first step in the evolutionary algorithm is producing the initial population, which will go through the process of evolution during a certain number of generations.
The evolutionary algorithm employed is a standard ($\mu$ + $\lambda$)\footnote{We set both to 100 empirically.} where mutation and crossover are applied to the current population to produce $\lambda$ offspring.
Individuals in the current population and their offspring are then evaluated by the algorithm to find the fittest $\mu$ number of individuals to survive to the next generation.
Once the specified number of generations (10, in our case) is reached, the evolutionary process ends and returns the final population.

\subsubsection{Initial Population}
The initial population consists of $\mu$ copies of the input movie title.
For each copy, a randomly selected noun, adjective or verb is replaced with a random word from the vocabulary.
We used \textit{Spacy}~\cite{spacy2} to parse titles. We inflect the substituting words using \textit{Pattern}~\cite{smedt2012pattern} to match the morphology of the original word when needed.
The altered titles assemble the initial population.

\subsubsection{Mutation and Crossover}
In our evolutionary algorithm, we implement one kind of mutation and crossover.
The mutation process substitutes words in the individual in the same fashion as done in the creation of the initial population.
The crossover employed is a standard single-point crossover, i.e. a random point in individuals is selected and words to the right of the point are switched between them.

\subsubsection{Evaluation}
In our evaluation metric, we propose four internal evaluation dimensions to measure the fitness of an individual.
These dimensions are (1)~prosody, (2)~semantic similarity to ``food'', (3)~semantic similarity to the original word, and (4)~number of altered words. 
The first two dimensions are maximized, whereas the last two are minimized. 

The prosody dimension is a weighted sum of four prosody sub-features, which are consonance, assonance, rhyme and alliteration.
This dimension measures the sound similarity between the original word and its substitution.
To measure the sound similarity, we use \textit{espeak-ng tool}\footnote{https://github.com/espeak-ng/espeak-ng} to generate IPA (international phonetic alphabet) transcriptions for assessing the prosody.

To measure the semantic similarity between two words, we employ a pre-trained \textit{Glove} model\footnote{https://nlp.stanford.edu/projects/glove/} with 6 billion tokens and a dimension size of 300. 
The model is trained on Wikipedia and English Gigaword Fifth Edition corpus.
Using the semantic model, the next dimension computes the maximum semantic similarity of words in the title to the word ``food''.

The third dimension measures the mean of the semantic similarity of new words to their original corresponding word.
We minimize this dimension to increase surprise, with the idea that a lower semantic similarity between the original word and its substitute would result in a bigger surprise.

The last dimension keeps track of the number of words modified in comparison to the original title.
Minimizing this dimension motivates that less substitutions are made to the title, which makes it more recognizable.

These are the criteria based on which the fitness of individuals is evaluated at the end of each generation to let only the best ones survive to the next generation.

\subsubsection{Selection and Filtering}
To reduce having a dominating dimension and motivate generating titles with diverse and balanced scores on all four dimensions, we opt for a non-dominant sorting algorithm --\textit{NSGA-II}-- \cite{DebNSGAII2002} as the selection algorithm.

During each iteration of the evolution, the current population and its offspring go through a filtering phase which filters out any duplicate titles.

\subsubsection{Final Verdict}
On top of individual evaluation metrics, we introduce master's final verdict, which is a way of telling whether the master likes the generated title.
The final verdict of the master is a binary decision, i.e. an individual is either good or not.
In practice, the final verdict is defined as conditional thresholds on each dimension.
These thresholds are 1) a positive non-zero value for prosody, 2) a positive non-zero  semantic similarity to ``food'', 3) a semantic similarity less than 0.5 of the new word to its original and 4) not more than 50 percent change of content words.

The master uses this functionality to express its liking to titles outside of its own creations such as those created by the apprentice. Whenever we talk about the master liking something in this paper, we mean that the final verdict has a Boolean value of true.

\subsection{Apprentice}

For the apprentices we use OpenNMT \cite{opennmt}, which implements an RNN based sequence to sequence model. The model has two RNN encoding layers and two RNN decoding layers.

The attention mechanism is the general global attention formulated by \cite{luong2015effective}. The difference to the OpenNMT default parameters in our system is that we use the copy attention mechanism which makes it possible for the model to copy words from the source. This is useful since the task is to translate within the same language.

All of the apprentice models described in this paper have been trained by using the same random seed to make their intercomparison possible.

\subsection{Different Parenting Styles}

We model computationally the four different parenting styles, authoritarian, authoritative, permissive and rejecting-neglecting, in the way the master interacts with the apprentice during the training process of the NMT model.

The training process is done iteratively. In each iteration, the apprentice is trained for 1000 training steps. After each iteration, the apprentice produces an output based on the 1276 popular IMDB movie titles. This output is then evaluated by the master accordingly to the parenting style in question and adjustments are made to the training data based on the master's parenting. The apprentices are trained for 20 iterations.

\subsubsection{The Authoritarian Master} only lets the apprentice learn from its own output. The apprentice is not exposed to any of the peer data and the apprentice's own creations are not taken into account.

\subsubsection{The Authoritative Master} lets the apprentice learn from its own creations and those peers who it considers good enough by the final verdict (this means 2446 titles). The apprentice can show its creations to the master after each training iteration, out of which the master picks the ones it likes and adds them to the training material of the apprentice. The training of the NMT model continues with the modified corpus.

\subsubsection{The Permissive Master} lets the apprentice learn from its own creations and all of the peer data. When the apprentice presents its own creations at the end of a training iteration, the master praises them all and adds them to the training data. 

\subsubsection{The Rejecting-Neglecting Master} does not care about the apprentice. The apprentice has no choice but to learn from its peers. The apprentice does not learn from its own creations because it receives no support from the master.

\subsection{Training the Apprentice}

The master is run once to create its own movie titles with food related puns. This parallel data of 8306 titles is shared across the different parenting styles. During the training process of the apprentice, the master does not generate new titles of its own, but only interferes in the selection of the parallel data used in the next training iteration as described in the sections above.

After each iteration, we calculate BLEU score \cite{papineni2002bleu} and a uni-gram PINC score \cite{chen2011collecting} for the outputs of the apprentices. We compare the outputs both to the training material coming from the master and the material from the peers. For each title generated by the apprentice, we take the maximum BLEU and minimum PINC score and take an average of them for each iteration.

BLEU score is traditionally used in machine translation to evaluate how good the final translation is in terms of a gold standard. We, however, do not use BLEU as a final evaluation metric, but rather use it to shed some light into how closely the outputs of the apprentices resemble those of the master or the peer written titles. BLEU measures the similarity, whereas PINC measures divergence from the original data. In other words, the higher the BLEU, the more closely the apprentice imitates and the higher the PINC the less it imitates the master or the peers.

\begin{figure}[!htb]
\center{\includegraphics[width=5cm]
{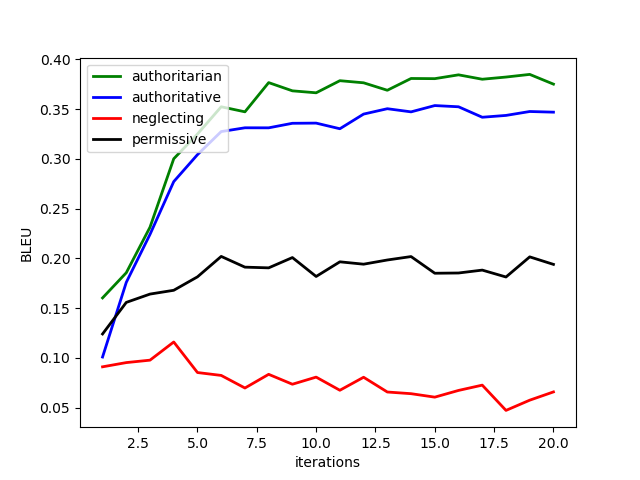}}
\caption{\label{fig:masterbleu} BLEU when comparing to the master}
\end{figure}

\begin{figure}[!htb]
\center{\includegraphics[width=5cm]
{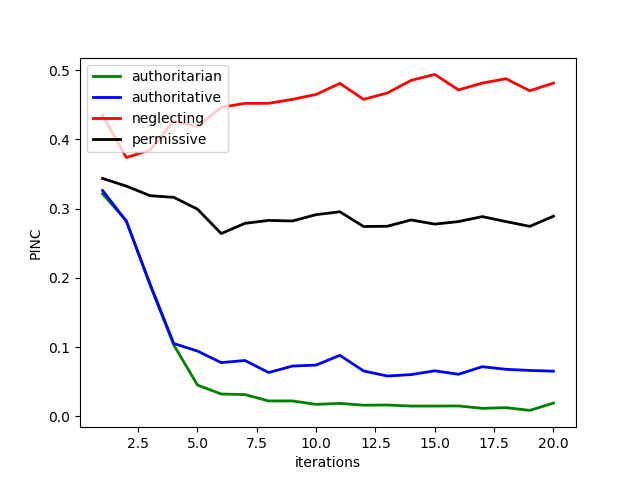}}
\caption{\label{fig:masterpinc} PINC when comparing to the master}
\end{figure}

As indicated by Figure \ref{fig:masterbleu}, the authoritarian scenario, where the training data consists only of the master's output, starts quickly producing the output most similar to the master. Where as the authoritative scenario leads to a bit less similarity to the master. The effect of the peer data is very well visible in the permissive and neglecting scenarios. The PINC scores in Figure \ref{fig:masterpinc} show the other side of the coin where the authoritative and authoritarian scenarios are the least divergent and the permissive and neglecting ones the most divergent.

\begin{figure}[!htb]
\center{\includegraphics[width=5cm]
{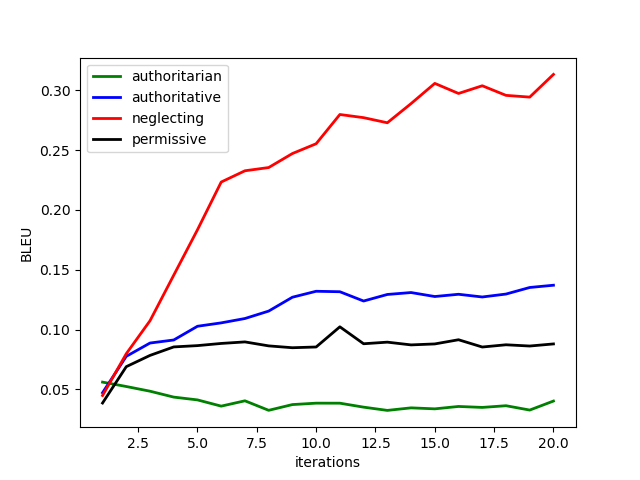}}
\caption{\label{fig:peerbleu} BLEU when comparing to peers}
\end{figure}

\begin{figure}[!htb]
\center{\includegraphics[width=5cm]
{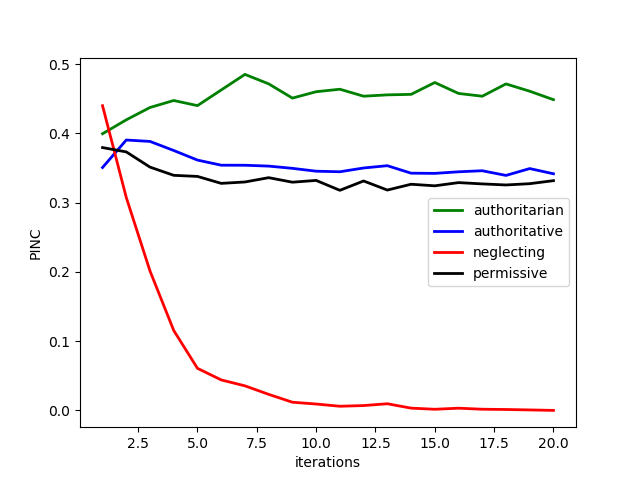}}
\caption{\label{fig:peerpinc} PINC when comparing to peers}
\end{figure}

When we do the BLEU comparison to the peer data as seen in Figure \ref{fig:peerbleu}, we can see that only the neglecting scenario leads to high similarity with the peers, where as the other scenarios are still quite low, the lowest being the authoritarian scenario. The PINC scores tell a similar story in Figure \ref{fig:peerpinc}, where the neglecting scenario leads to the least amount of divergence, leaving the authoritarian scenario the most divergent.

\section{Results and Evaluation}

In this section, we show some of the results produced by the different systems. In addition, we evaluate the different parenting style scenarios after each iteration with the master's appreciation function. Later, an evaluation is conducted by humans.

\subsection{Results and Master's Liking}

Results from the approaches can be seen in Table \ref{examples}. The master did not produce any training data for the last two titles in the examples. Looking at these results qualitatively, in broad lines, the permissive and neglecting scenarios produced worse output than the ones exposed to the master's training data. The apprentice exposed to authoritarian parenting struggles in producing output for titles not present in the training data. The authoritative scenario leads to the most consistent results. The quantitative human evaluation in the next section is used to verify these initial observations.

\begin{table*}[t]
\centering
\resizebox{1.0\textwidth}{!}{%
{\large
\begin{tabular}{|l|l|l|l|l|l|}
\hline
\textbf{original} & \textbf{master} & \textbf{authoritarian} & \textbf{authoritative} & \textbf{permissive} & \textbf{neglecting} \\ \hline
the butterfly effect & \textit{the brewery effect} & \textit{the butterfly kimchi} & \textit{the butterfly chicken} & \textit{the butterfly effect} & \textit{\begin{tabular}[c]{@{}l@{}}the lasagna \\ effect lazarus\end{tabular}} \\ \hline
\begin{tabular}[c]{@{}l@{}}how to train \\ your dragon\end{tabular} & \textit{\begin{tabular}[c]{@{}l@{}}how to train \\ your pepperoni\end{tabular}} & \textit{\begin{tabular}[c]{@{}l@{}}how to train \\ your avocado\end{tabular}} & \textit{\begin{tabular}[c]{@{}l@{}}how to train \\ your pepperoni\end{tabular}} & \textit{\begin{tabular}[c]{@{}l@{}}how to train \\ your bacon\end{tabular}} & \textit{\begin{tabular}[c]{@{}l@{}}how to train \\ your bacon\end{tabular}} \\ \hline
\begin{tabular}[c]{@{}l@{}}fantastic beasts and\\ where to find them\end{tabular} & \textit{----} & \textit{\begin{tabular}[c]{@{}l@{}}fantastic ordinary \\ and where to find\end{tabular}} & \textit{\begin{tabular}[c]{@{}l@{}}fantastic beets and \\ where to find them\end{tabular}} & \textit{\begin{tabular}[c]{@{}l@{}}fantastic beefs and \\ where to find them\end{tabular}} & \textit{\begin{tabular}[c]{@{}l@{}}fantastic beets and \\ where to find them\end{tabular}} \\ \hline
under the skin & \textit{----} & \textit{under the cereals} & \textit{under the silver cake} & \textit{under the 13th} & \textit{fryday the 13} \\ \hline
\end{tabular}
}}
\caption{Examples of the final output of the different models}
\label{examples}
\end{table*}

\begin{figure}[!htb]
\center{\includegraphics[width=5cm]
{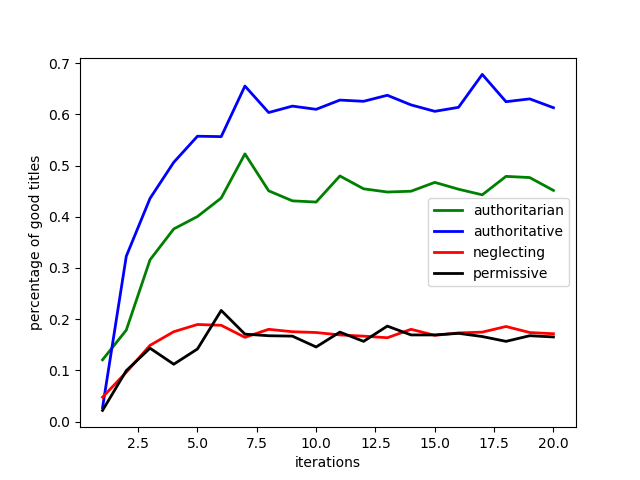}}
\caption{\label{fig:masterlikes} Master's liking of the output}
\end{figure}

Another way to look at the results is to use the appreciation metrics implemented in the master. Figure \ref{fig:masterlikes} shows the percentage of how many titles the master liked after each training iteration. 

As we can see, the appreciation the master has ranks the authoritarian and authoritative scenarios higher than the permissive and neglecting ones. Even in the authoritarian case, the master does not like all of the output produced by the apprentice, which shows that the appreciation learned by the apprentices is different from the one implemented in the master.

It is interesting to see to what extent the master's liking correlates with the evaluation results of the human judges. This can reveal more information about the adequacy of the appreciation of the master in this creative task. Or does the master's appreciation only tell about obedience when applied to the apprentices' output?

\subsection{Evaluation Questions}

In this section we provide some reasoning in our selection of the evaluation questions that are presented to the human judges. Earlier, we defined the creativity in the case of pun generation using the creative tripod as our theoretical framework. This means that on a higher level, our evaluation questions should evaluate \textit{skill}, \textit{appreciation} and \textit{imagination}.

\subsubsection{Skill} Our definition for skill stated that the system should be able to take an existing movie title and produce a food related pun as an output. A further requirement was that the original title should be recognizable from the generated one.

\begin{enumerate}
  \item The title has a pun in it
  \item The title is related to food
  \item The original title is recognizable
\end{enumerate}

The evaluation questions described above are designed to evaluate the requirements set for skill. We evaluate whether a pun is perceived and whether the new title relates to food separately, as it might be that the replacement word delivers a pun, but is not food related or vice-versa.

\subsubsection{Appreciation} We defined appreciation from the humor stand point. A good title with a pun is also funny. For something to be funny, i.e. humorous, the pun has to exhibit coherence and surprise.

\begin{enumerate}
  \setcounter{enumi}{3}
  \item The title is humorous
  \item The pun is surprising
  \item The pun makes sense in the context of the original movie
\end{enumerate}

We choose to evaluate the overall humor value of the title separately from the components that constitute it. The last two questions are designed to evaluate surprise and coherence respectively.

\subsubsection{Imagination} We used Boden's dichotomy to establish the definition of imagination. The minimal requirement was set to P-creativity. However, P-creativity can easily be verified by looking at the training data and the final output, if the output is different from the training material, there is P-creativity. Therefore, we use human judges to assess the H-creativity of the outputs.

\begin{enumerate}
  \setcounter{enumi}{6}
  \item The pun in the title is obvious
  \item The pun in the title sounds familiar
\end{enumerate}

If the pun is obvious, it probably is not too H-creative, as an obvious pun could be said by just about anyone, also if the pun sounds familiar, it has probably been said by someone before.

\subsection{Human Evaluation}

We take a random sample of 20 original movie titles that were only present in the training data provided by the master, 20 titles that were only present in the peer data and 20 titles that were in both sources of parallel data. We evaluate the creative output of each apprentice for these randomly sampled titles. In addition, we evaluate the master's output for the 40 titles of the sample it had generated movie title puns for. As the master has generated multiple creative titles per original title, we pick one randomly for each original title. Altogether, we are evaluating 280 computer created titles.

\begin{table*}[t]
\centering
\resizebox{1.0\textwidth}{!}{%
{\large
\begin{tabular}{|l|l|l|l|l|l|l|l|l|l|l|l|l|l|l|l|l|l|} 
\hline
\textbf{Style}  & \textbf{Training}  & \multicolumn{2}{c|}{\textbf{Q1} } & \multicolumn{2}{c|}{\textbf{Q2} } & \multicolumn{2}{c|}{\textbf{Q3} } & \multicolumn{2}{c|}{\textbf{Q4} } & \multicolumn{2}{c|}{\textbf{Q5} } & \multicolumn{2}{c|}{\textbf{Q6} } & \multicolumn{2}{c|}{\textbf{Q7} } & \multicolumn{2}{c|}{\textbf{Q8} }         \\ 
\hline
                &                    & $\mu_{x}$               & $SD$    & $\mu_{x}$               & $SD$    & $\mu_{x}$               & $SD$    & $\mu_{x}$               & $SD$    & $\mu_{x}$               & $SD$   & $\mu_{x}$               & $SD$   & $\mu_{x}$               & $SD$   & $\mu_{x}$               & $SD$            \\ 
\hline
authoritarian   & both               & \textbf{3.35 }          & 1.08    & 2.65                    & 1.26    & 3.33                    & 1.06    & \textbf{3.02 }          & 1.10    & \textbf{3.08 }          & 0.99   & 3.03                    & 1.01   & 3.16                    & 0.99   & 3.10                    & 1.03            \\
authoritarian   & peer only          & 2.97                    & 1.18    & 2.14                    & 1.17    & \textbf{3.44 }          & 1.13    & 2.66                    & 1.15    & 2.82                    & 1.08   & \textbf{3.10 }          & 1.11   & \textbf{3.03 }          & 1.12   & 3.11                    & 1.13            \\
authoritarian   & master only        & 3.34                    & 1.07    & \textbf{2.89 }          & 1.28    & 3.30                    & 1.05    & 3.00                    & 1.12    & 3.07                    & 1.05   & 3.04                    & 1.02   & 3.12                    & 1.02   & \textbf{3.07 }          & 1.04            \\ 
\hline
authoritative   & both               & 3.43                    & 1.08    & 3.09                    & 1.31    & 3.28                    & 1.12    & 3.08                    & 1.16    & 3.08                    & 1.05   & 3.02                    & 1.05   & \textbf{3.22}                    & 1.06   & \textbf{3.13 }          & 1.05   \\
authoritative   & peer only          & 3.41                    & 1.14    & \textbf{3.23 }          & 1.35    & 3.37                    & 1.21    & 3.13                    & 1.16    & 3.07                    & 1.08   & 3.08                    & 1.09   & 3.24                    & 1.10   & 3.16                    & 1.13            \\
authoritative   & master only        & \textbf{\textit{3.47 }} & 1.03    & 3.15                    & 1.33    & \textbf{3.41 }          & 1.08    & \textbf{3.17 }          & 1.11    & \textbf{\textit{3.16 }} & 1.02   & \textbf{3.16 }          & 1.04   & 3.28           & 1.01   & 3.24                    & 1.06            \\ 
\hline
master          & both               & 3.38                    & 1.07    & \textbf{2.79 }          & 1.28    & 3.29                    & 1.06    & 3.07                    & 1.12    & \textbf{3.09 }          & 1.04   & \textbf{3.10 }          & 1.06   & 3.22                    & 1.02   & 3.14                    & 1.05   \\
master          & master only        & \textbf{3.40 }          & 1.07    & 2.61                    & 1.30    & \textbf{3.34 }          & 1.11    & \textbf{3.10 }          & 1.15    & \textbf{3.09 }          & 1.02   & 3.04                    & 1.03   & \textbf{3.17 }          & 1.04   & \textbf{3.11 }          & 1.06            \\ 
\hline
neglecting      & both               & \textbf{3.45 }          & 1.11    & \textbf{\textit{3.28 }} & 1.32    & \textbf{3.34 }          & 1.11    & \textbf{\textit{3.28 }} & 1.12    & \textbf{\textit{3.16 }} & 1.06   & 3.12                    & 1.04   & 3.21                    & 1.08   & 3.22                    & 1.07            \\
neglecting      & peer only          & 3.36                    & 1.07    & 3.02                    & 1.37    & 3.31                    & 1.11    & 3.12                    & 1.15    & 3.09                    & 1.02   & \textbf{3.14 }          & 1.04   & 3.19                    & 1.02   & \textbf{3.14 }          & 1.06            \\
neglecting      & master only        & 3.28                    & 1.13    & 2.87                    & 1.35    & \textbf{3.34 }          & 1.12    & 3.09                    & 1.14    & 3.05                    & 1.05   & 3.07                    & 1.06   & \textbf{3.15 }          & 1.06   & 3.18                    & 1.08            \\ 
\hline
permissive      & both               & \textbf{3.23 }          & 1.18    & 2.67                    & 1.38    & 3.59                    & 1.06    & \textbf{3.06 }          & 1.13    & \textbf{3.08 }          & 1.08   & \textbf{\textit{3.30 }} & 1.04   & 3.21                    & 1.07   & 3.30                    & 1.10            \\
permissive      & peer only          & 3.05                    & 1.19    & \textbf{2.87 }          & 1.39    & 3.25                    & 1.18    & 2.88                    & 1.13    & 2.88                    & 1.09   & 3.00                    & 1.08   & 2.99                    & 1.11   & 3.04                    & 1.14            \\
permissive      & master only        & 3.09                    & 1.23    & 2.32                    & 1.24    & \textbf{\textit{3.64 }} & 1.11    & 2.88                    & 1.15    & 2.91                    & 1.12   & 3.07                    & 1.15   & \textbf{\textit{2.98 }} & 1.13   & \textbf{\textit{3.04 }} & 1.12            \\
\hline
\end{tabular}
}}
\caption{Mean and standard deviation.}
\label{tab:mean-std}
\end{table*}

\begin{table*}[t]
\centering
\resizebox{0.8\textwidth}{!}{%
{\small
\begin{tabular}{|l|l|l|l|l|l|l|l|l|} 
\hline
 \textbf{Style}  & \textbf{Q1}       & \textbf{Q2}       & \textbf{Q3}       & \textbf{Q4}       & \textbf{Q5}       & \textbf{Q6}       & \textbf{Q7}       & \textbf{Q8}       \\ 
\hline
authoritarian    & 78.33\%           & 31.67\%           & 83.33\%           & 31.67\%           & 48.33\%           & 68.33\%           & 70.00\%           & 68.33\%           \\
authoritative    & \textbf{93.33\% } & \textbf{60.00\% } & 83.33\%           & 56.67\%           & 63.33\%           & 66.67\%           & 90.00\%           & 70.00\%           \\
master           & 92.50\%           & 37.50\%           & \textbf{87.50\% } & 62.50\%           & 60.00\%           & 65.00\%           & 80.00\%           & \textbf{67.50\%}  \\
neglecting       & 86.67\%           & \textbf{60.00\% } & 81.67\%           & \textbf{66.67\% } & \textbf{65.00\% } & \textbf{73.33\% } & 75.00\%           & 78.33\%           \\
permissive       & 66.67\%           & 33.33\%           & 85.00\%           & 43.33\%           & 46.67\%           & \textbf{73.33\% } & \textbf{68.33\% } & 71.67\%           \\
\hline
\end{tabular}
}}
\caption{Percentage of movie titles having an average score by judges greater than 3}
\label{tab:percentage}
\end{table*}

The evaluation was conducted on a crowd-sourcing platform called Figure Eight\footnote{https://www.figure-eight.com/}. The platform assigned people to conduct evaluation in such a way that each title was evaluated by 35 different users. The users could choose how many titles they wanted to evaluate. The results of the evaluation are show in Table \ref{tab:mean-std}. In the Training column, \textit{both}, \textit{peer only} and \textit{master only} indicate whether the original title was only present in the master produced training data, peer produced training data or in both respectively.

The authoritarian scenario didn't get the best average score for any of the test questions and neither did the master. They both score particularly low on the Q2, which reflects the fact that some of the words in the HTOED food and drink taxonomy were only loosely related to food such as \textit{steam} and \textit{spit}. It is interesting to note that the authoritarian scenario gets the best results for Q3, Q6 and Q7 for titles it did not encounter in the training data, in other words it has developed an appreciation of its own that does not just mimic what the master produces and fail otherwise. In light of these results, we can deduce that the master produced worse titles with food puns than real people, which left both the master and the authoritarian scenario without the first place on any of the test questions.

The authoritative scenario, which was the highest ranking one according to master's liking as seen in Figure \ref{fig:masterlikes}, got the best results for Q1 and Q5. This means that it succeeds the best in the main task of generating puns and they end up being the most surprising ones. It is also the only one that produces consistently good results (above 3 on the average) for all training test sets for Q1-Q6, unfortunately the results for Q7 and Q8 are also above 3 on the average meaning that it does not rank high on H-creativity.

The same consistency can not be perceived in the the permissive case as scores below 3 are common across the test questions. It however, manages to score the best for Q3 and Q7-Q8, in other words, it can achieve the best H-creativity and the original titles can be the most recognizable, although not consistently so. This shows, that even though the appreciation the master has might not be spot on, as it is not able to produce the best scoring titles, having moderation on the peer data and critical assessment of the apprentice generated results during the training by the master, has a positive effect on the consistency of the results. In the permissive scenario, the apprentice was exposed to everything without criticism and in the authoritative some criticism was used to filer the training data, which made the authoritative scenario more consistent, but less H-creative.

Finally, the neglecting scenario gets the best scores for the Q2, Q4 and Q5. It is the best one at producing humorous, surprising and food related titles. It is quite consistent with only the results for Q2 with previously unseen titles giving a score that is inferior to 3. The good results of the neglecting scenario serve as an additional proof to the fact that the output of the master is worse than human written titles.

Table \ref{tab:percentage} shows the results form another stand point. The table shows overall how many titles got the average rating above 3 for each test question. These numbers are in line with what was previously discussed about the Table \ref{tab:mean-std}. The authoritarian scenario leads to the worst performance, but this time master gets the highest percentage point of titles above 3 for Q3. In the authoritative scenario most of the titles have a clear pun and are related to food with the highest percentage point. The permissive scenario holds the best percentage points for Q6 and Q7. And the neglecting gets the best percentage points in Q2, Q4, Q5 and Q6.

\section{Discussion and Future Work}

The evaluation results were not completely in line with what we can observe by looking at the titles output by the different methods by ourselves. This raises the question whether our definition for creativity in movie title puns is adequate and whether the evaluation questions we formulated based on the definition really measure what they were designed to measure. Because we have worked with a clear definition for creativity in this paper, it is possible to take this under a critical study in the future. We also find evident that qualitative research on the output titles with respect to the quantitative results we got from the human judges is needed to evaluate the evaluation itself.

Having a master with appreciation filter the parallel data of the apprentice was beneficial for consistency (see authoritative vs permissive). Although the evaluation results showed that the appreciation is not in par with that of a real human, the implication remains that a good external appreciation can be beneficial for the learning outcome of the apprentice model. As we used a rather generic NMT model for the apprentice, our findings might be of a use in more traditional context of sequence-to-sequence models such as machine translation, text summarization or paraphrasing.

For now, the master and apprentice have been studied in a social vacuum, where peer data is the only link to the surrounding world. However, in the future it would be fruitful to see how the creative outcome changes when the master and the apprentice are exposed to a more complex social system such as the one described by Bronfenbrenner \cite{bronfenbrenner1979ecology}. In such a society, the master would also be under a social pressure in changing its own standards of appreciation.

\section{Conclusions}

This work has presented one of the first contributions to the field of computational social creativity where the computationally creative agents are in a hierarchical social relation. This asymmetry offers an intriguing setting for studying socialization of computational agents from the creativity perspective.

Despite building our definition of creativity upon an existing theory and formulating the test questions based on the definition, the quantitative evaluation left many questions unanswered. The results presented in this paper call for qualitative evaluation to understand the phenomenon of evaluation in this particular context.

Nevertheless, our findings suggest that having appreciation in parenting, or training, an NMT model can be of a benefit. The applicability of these finding into sequence-to-sequence deep learning models in a more generalized fashion is an interesting research question on its own right.

\bibliographystyle{iccc}
\bibliography{iccc}

\end{document}